\documentclass[10pt, a4paper]{article}
\pdfoutput=1
\usepackage{lrec}
\usepackage{graphicx}
\usepackage{tabularx}
\usepackage{soul}
\usepackage[gen]{eurosym}

\usepackage{epstopdf}
\usepackage[utf8]{inputenc}
\usepackage[T1]{fontenc}
\usepackage[draft]{hyperref}
\usepackage{booktabs}
\usepackage{amsmath}
\usepackage{url}
\usepackage{microtype}
\usepackage{xcolor}

\usepackage{scalefnt}

\usepackage{titlesec}
\titleformat{\section}{\normalfont\large\bf\center}{\thesection.}{1em}{}
\titleformat{\subsection}{\normalfont\SmallTitleFont\bf\raggedright}{\thesubsection.}{1em}{}
\titleformat{\subsubsection}{\normalfont\normalsize\bf\raggedright}{\thesubsubsection.}{1em}{}
\renewcommand\thesection{\arabic{section}}
\renewcommand\thesubsection{\thesection.\arabic{subsection}}
\renewcommand\thesubsubsection{\thesubsection.\arabic{subsubsection}}

\title{PO-EMO: Conceptualization, Annotation, and Modeling of\\ Aesthetic Emotions in German and English Poetry}

\name{Thomas Haider$^{1,3}$, Steffen Eger$^2$, Evgeny Kim$^3$, Roman Klinger$^3$, Winfried Menninghaus$^1$}

\address{%
  $^{1}$Department of Language and Literature, Max Planck Institute for Empirical Aesthetics\\
  $^{2}$NLLG, Department of Computer Science, Technische Universitat Darmstadt\\
  $^{3}$Institut f\"ur Maschinelle Sprachverarbeitung, University of Stuttgart \\
  \{thomas.haider, w.m\}@ae.mpg.de, eger@aiphes.tu-darmstadt.de\\
  \{roman.klinger, evgeny.kim\}@ims.uni-stuttgart.de\\}

\abstract{Most approaches to emotion analysis of social media,
  literature, news, and other domains focus exclusively on basic
  emotion categories as defined by Ekman or Plutchik. However, art
  (such as literature) enables engagement in a broader range of more
  complex and subtle emotions. These have been shown to also include
  mixed emotional responses. We consider emotions in poetry as they are
  \textit{elicited in the reader}, rather than what is
  \textit{expressed in the text} or \textit{intended by the
    author}. Thus, we conceptualize a set of \textit{aesthetic
    emotions} that are predictive of aesthetic appreciation in the
  reader, and allow the annotation of multiple labels per line to
  capture mixed emotions within their context. We evaluate this novel
  setting in an annotation experiment both with carefully trained
  experts and via crowdsourcing. Our annotation with experts leads to
  an acceptable agreement of $\kappa=.70$, resulting in a consistent
  dataset for future large scale analysis. Finally, we conduct first
  emotion classification experiments based on BERT, showing that
  identifying aesthetic emotions is challenging in our data, with up
  to .52 F1-micro on the German subset. Data and resources are
  available
  at \url{https://github.com/tnhaider/poetry-emotion}. \\
  \newline \Keywords{Emotion, Aesthetic Emotions, Literature, Poetry,
    Annotation, Corpora, Emotion Recognition, Multi-Label} }

\begin{document}

\maketitleabstract

\section{Introduction}
\label{sec:introduction}
Emotions are central to human experience, creativity and
behavior. Models of affect and emotion, both in psychology and natural
language processing, commonly operate on predefined categories,
designated either by \textit{continuous scales} of, e.g.,\
\textit{Valence}, \textit{Arousal} and \textit{Dominance}
\cite{mohammad2016sentiment} or \textit{discrete emotion labels}
(which can also vary in intensity). Discrete sets of emotions often
have been motivated by theories of basic emotions, as proposed by
\newcite{Ekman1992}---\textit{Anger}, \textit{Fear}, \textit{Joy},
\textit{Disgust}, \textit{Surprise}, \textit{Sadness}---and
\newcite{Plutchik1991}, who added \textit{Trust} and
\textit{Anti\-cipation}. These categories are likely to have evolved as they motivate behavior that is directly relevant for
survival. However, \emph{art reception} typically
presupposes a situation of safety and therefore offers special
opportunities to engage in a broader range of more complex and subtle
emotions. These differences between real-life and art contexts have
not been considered in natural language processing work so far.

To emotionally move readers is considered a prime goal of literature since Latin antiquity
\cite{Johnson2016,menninghaus2019aesthetic,menninghaus2015towards}. Deeply moved readers shed tears or get
chills and goosebumps even in lab settings
\cite{wassiliwizky2017tears}. In cases like these, the emotional
response actually implies an aesthetic evaluation: narratives that
have the capacity to move readers are evaluated as good and powerful
texts for this very reason. Similarly, feelings of suspense
experienced in narratives not only respond to the trajectory of
the plot's content, but are also directly predictive of aesthetic
liking (or disliking). Emotions that exhibit this dual capacity have
been defined as ``aesthetic emotions''
\cite{menninghaus2019aesthetic}.
Contrary to the negativity bias of classical emotion catalogues,
emotion terms used for aesthetic evaluation purposes include far more
positive than negative emotions. At the same time, many
overall positive aesthetic emotions encompass negative or mixed
emotional ingredients \cite{menninghaus2019aesthetic}, e.g.,
feelings of suspense include both hopeful and fearful
anticipations.

For these reasons, we argue that the analysis of
literature (with a focus on poetry) should rely on specifically
selected emotion items rather than on the narrow range of basic
emotions only. Our selection is based on previous research on this
issue in psychological studies on art reception and, specifically, on
poetry. For instance, \newcite{knoop2016mapping} found that
\textit{Beauty} is a major factor in poetry reception.

We primarily adopt and adapt emotion terms that
\newcite{schindler2017measuring} have identified as aesthetic emotions in their study on how to measure and categorize such particular affective states. Further, we consider the aspect that, when selecting specific
emotion labels, the perspective of annotators plays a major
role. Whether emotions are \textit{elicited in the reader},
\textit{expressed in the text}, or \textit{intended by the author}
largely changes the permissible labels. For example, feelings of \textit{Disgust} or \textit{Love} might be intended or expressed in the text, but the text might still fail to elicit corresponding feelings as these concepts presume a strong reaction in the reader. Our focus here was on the actual emotional experience of the readers rather than on hypothetical intentions of authors. We opted for this reader perspective based on previous research in NLP
\cite{buechel-hahn-2017-emobank,Buechel2017} and work in empirical
aesthetics \cite{menninghaus2017emotional}, that specifically measured the reception of poetry. Our final set of emotion labels consists of \textit{Beauty/Joy, Sadness,
  Uneasiness, Vitality/Energy, Suspense, Awe/Sublime, Humor, Annoyance}, and
\textit{Nostalgia}.\footnote{The concepts \textit{Beauty} and \textit{Awe/Sublime} primarily define object-based aesthetic virtues. \newcite{kant2000critique} emphasized that such virtues are typically intuitively felt rather than rationally computed. Such \textit{feelings} of \textit{Beauty} and \textit{Sublime} have therefore come to be subsumed under the rubrique of \textit{aesthetic emotions} in recent psychological research \cite{menninghaus2019aesthetic}. For this reason, we refer to the whole set of category labels as \textit{emotions} throughout this paper.}

In addition to selecting an adapted set of emotions, the annotation of poetry brings further challenges, one of which is the choice of the
appropriate unit of annotation. Previous work considers words\footnote{to create emotion dictionaries}
\cite{Mohammad2013,Strapparava2004},
sentences \cite{Alm2005,Aman2007}, utterances \cite{Cevher2019},
sentence triples \cite{Kim2018}, or paragraphs \cite{Liu2019} as the units of annotation.  For poetry, reasonable units follow the logical document structure of poems, i.e., verse (line), stanza, and, owing to its relative shortness, the
complete text. The more coarse-grained the unit, the more difficult
the annotation is likely to be, but the more it may also enable the
annotation of emotions in context. We find that annotating
fine-grained units (lines) that are hierarchically ordered within a larger context
(stanza, poem) caters to the specific structure of poems, where
emotions are regularly mixed and are more
interpretable within the whole poem. Consequently, we allow the mixing of emotions already at line level through multi-label annotation.

The remainder of this paper includes (1) a report of the annotation
process that takes these challenges into consideration, (2) a
description of our annotated corpora, and (3) an implementation of
baseline models for the novel task of aesthetic emotion annotation in
poetry. In a first study, the annotators work on the annotations in a
closely supervised fashion, carefully reading each verse, stanza, and
poem. In a second study, the annotations are performed via
crowdsourcing within relatively short time periods with annotators not
seeing the entire poem while reading the stanza. Using these two
settings, we aim at obtaining a better understanding of the advantages
and disadvantages of an expert vs.\ crowdsourcing setting in this
novel annotation task. Particularly, we are interested in estimating the
potential of a crowdsourcing environment for the task of self-perceived
emotion annotation in poetry, given time and cost overhead associated
with in-house annotation process (that usually involve training and
close supervision of the annotators).

We provide the final datasets of German and English language poems
annotated with reader emotions on verse level at
\url{https://github.com/tnhaider/poetry-emotion}.

\section{Related Work}
\subsection{Poetry in Natural Language Processing}
Natural language understanding research on poetry has investigated 
\textit{stylistic variation}
\cite{kaplan2007computational,kao2015computational,voigt2013tradition},
with a focus on broadly accepted formal features such as
\textit{meter}
\cite{greene2010automatic,agirrezabal-etal-2016-machine,estes2016supervised}
and \textit{rhyme} \cite{reddy2011unsupervised,haider2018supervised},
as well as \textit{enjambement} \cite{ruiz2017enjambment,baumann2018style} and
\textit{metaphor} \cite{kesarwani2017metaphor,reinig2019metaphor}.  Recent work
has also explored the relationship of poetry and prose, mainly on a
syntactic level
\cite{krishna2019poetry,gopidi2019computational}. Furthermore, poetry
also lends itself well to semantic (change) analysis
\cite{haider2019diachronic,haider2019semantic}, as linguistic
invention \cite{underwood2012,herbelot2014semantics} and succinctness
\cite{roberts2000poetry} are at the core of poetic production.

Corpus-based analysis of emotions in poetry has been considered, but
there is no work on German, and little on English.
\newcite{kao2015computational} analyze English poems with word
associations from the Harvard Inquirer and LIWC, within the categories
\textit{positive/negative outlook}, \textit{positive/negative emotion}
and \textit{phys./psych.\
  well-being}. \newcite{hou-frank-2015-analyzing} examine the binary
sentiment polarity of Chinese poems with a weighted personalized
PageRank algorithm.  \newcite{barros2013automatic} followed a tagging
approach with a thesaurus to annotate words that are similar to the
words `Joy', `Anger', `Fear' and `Sadness' (moreover translating these
from English to Spanish). With these word lists, they distinguish the
categories `Love', `Songs to Lisi', `Satire' and
`Philosophical-Moral-Religious' in Quevedo's poetry. Similarly,
\newcite{alsharif2013emotion} classify unique Arabic `emotional text
forms' based on word unigrams.

\newcite{Mohanty2018} create a corpus of 788 poems in the Indian Odia
language, annotate it on text (poem) level with binary negative and
positive sentiment, and are able to distinguish these with moderate
success.  \newcite{Sreeja2019} construct a corpus of 736 Indian
language poems and annotate the texts on Ekman's six categories + Love
+ Courage. They achieve a Fleiss Kappa of .48.

In contrast to our work, these studies focus on basic emotions and
binary sentiment polarity only, rather than addressing aesthetic
emotions. Moreover, they annotate on the level of complete poems
(instead of fine-grained verse and stanza-level).

\subsection{Emotion Annotation}
Emotion corpora have been created for different tasks and with
different annotation strategies, with different units of analysis and
different foci of emotion perspective (reader, writer, text). Examples
include the ISEAR dataset \cite{scherer1994evidence} (document-level);
emotion annotation in children stories \cite{Alm2005} and news
headlines \cite{strapparava2007semeval} (sentence-level); and
fine-grained emotion annotation in literature by \newcite{Kim2018}
(phrase- and word-level). We refer the interested reader to an
overview paper on existing corpora \cite{Bostan2018}.

We are only aware of a limited number of publications which look in
more depth into the emotion
perspective. \newcite{buechel-hahn-2017-emobank} report on an
annotation study that focuses both on writer's and reader's emotions
associated with English sentences. The results show that the reader
perspective yields better inter-annotator agreement.
\newcite{Yang2009} also study the difference between writer and reader
emotions, but not with a modeling perspective. The authors find that
positive reader emotions tend to be linked to positive writer emotions
in online blogs.

\subsection{Emotion Classification}
The task of emotion classification has been tackled before using
rule-based and machine learning approaches. Rule-based emotion
classification typically relies on lexical resources of
emotionally charged words
\cite{Strapparava2004,Esuli2006,Mohammad2013} and offers a
straightforward and transparent way to detect emotions
in text. 

In contrast to rule-based approaches, current models for emotion
classification are often based on neural networks and commonly use
word embeddings as features. \newcite{Schuff2017} applied models from
the classes of CNN, Bi\-LSTM, and LSTM and compare them to linear
classifiers (SVM and MaxEnt), where the BiLSTM shows best results with
the most balanced precision and recall. \newcite{AbdulMageed2017}
claim the highest F$_1$ with gated recurrent unit networks
\cite{Chung2015} for Plutchik's emotion model.  More recently, shared
tasks on emotion analysis \cite{Mohammad2018,klinger-etal-2018-iest}
triggered a set of more advanced deep learning approaches, including
BERT \cite{Devlin2019} and other transfer learning methods
\cite{dankers-etal-2019-modelling}.

\section{Data Collection}

\begin{figure}
\centering
\includegraphics{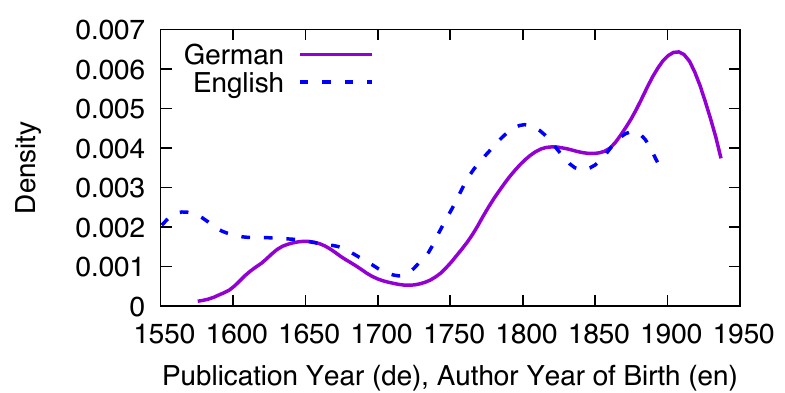}
\caption{Temporal distribution of poetry corpora (Kernel Density Plots
  with bandwidth = 0.2).}
\label{fig:temporal.density}
\end{figure}
\begin{table}
    \centering
    \begin{tabular}{lrr} 
        \toprule 
         & German & English  \\ 
         \cmidrule(r){1-1}\cmidrule(rl){2-2}\cmidrule(l){3-3}
         \# tokens & 20403 & 8082 \\
         \# lines & 3650 & 1240 \\
         \# stanzas & 731 & 174 \\
         \# poems & 158 & 64 \\
         \# authors & 51 & 22 \\
         \bottomrule
    \end{tabular}
    \caption{Statistics on our poetry corpora \emph{PO-EMO}.\\ Tokens without punctuation.}
    \label{table:data_statistics}
\end{table}

%GERMAN
%lines
%1198 in gold
%2452 in tsv
%== 3650
% old 3651

%tokens
%6559 in gold
%13844 in tsv
%== 20403
% old: 20647

For our annotation and modeling studies, we build on top of two
poetry corpora (in English and German), which we refer to as
\emph{PO-EMO}. This collection represents important contributions to
the literary canon over the last 400 years. We make this resource
available in TEI P5
XML\footnote{\url{https://tei-c.org/guidelines/p5/}} and an
easy-to-use tab separated format.
Table \ref{table:data_statistics} shows a size overview of these data
sets. Figure~\ref{fig:temporal.density} shows the distribution of our
data over time via density plots. Note that both corpora show a
relative underrepresentation before the onset of the romantic period
(around 1750).

\subsection{German}
The German corpus contains poems available from the website
\url{lyrik.antikoerperchen.de} (ANTI-K), which provides a platform for
students to upload essays about poems. The data was available in the
Hypertext Markup Language, with clean line and stanza
segmentation, which we transformed into TEI P5. ANTI-K also has extensive metadata, including author
names, years of publication, numbers of sentences, poetic genres, and
literary periods, that enable us to gauge the distribution of poems
according to periods. The 158 poems we consider (731 stanzas) are
dispersed over 51 authors and the New High German timeline (1575--1936
A.D.). This data has been annotated, besides emotions, for meter,
rhythm, and rhyme in other studies \cite{haider2018supervised,haider2020rhythm}.

\subsection{English}
The English corpus contains 64 poems of popular English writers. It
was partly collected from Project Gutenberg with the GutenTag
tool,\footnote{\url{https://gutentag.sdsu.edu/}} and, in addition,
includes a number of hand selected poems from the modern period and
represents a cross section of popular English poets. We took care to
include a number of female authors, who would have been
underrepresented in a uniform sample. Time stamps in the corpus are
organized by the birth year of the author, as assigned in Project
Gutenberg.

\section{Expert Annotation}
In the following, we will explain how we compiled and annotated three
data subsets, namely, (1) 48 German poems with gold annotation. These
were originally annotated by three annotators. The labels were then
aggregated with majority voting and based on discussions among the
annotators. Finally, they were curated to only include one gold
annotation. (2) The remaining 110 German poems that are used to compute the agreement in table \ref{table:expert_agreements} and (3) 64 English
poems contain the raw annotation from two annotators.

We report the genesis of our annotation guidelines including the
emotion classes. With the intention to provide a language resource for
the computational analysis of emotion in poetry, we aimed at
maximizing the consistency of our annotation, while doing justice to
the diversity of poetry. We iteratively improved the guidelines and
the annotation workflow by annotating in batches, cleaning the class
set, and the compilation of a gold standard. The final overall cost of
producing this expert annotated dataset amounts to approximately
\euro{3,500}.

\subsection{Workflow}
The annotation process was initially conducted by three female
university students majoring in linguistics and/or literary studies,
which we refer to as our ``expert annotators''. We used the INCePTION
platform for
annotation\footnote{\url{https://www.informatik.tu-darmstadt.de/ukp/research_6/current_projects/inception/index.en.jsp}}
\cite{klie2018inception}.
Starting with the German poems, we annotated in batches of about 16 (and later in some cases 32) poems. After each batch, we computed agreement statistics including heatmaps, and provided this feedback to the annotators.  For
the first three batches, the three annotators produced a gold standard using a majority vote for each line. Where this was inconclusive, they developed an adjudicated annotation based on
discussion. Where necessary, we encouraged the annotators to aim for more consistency, as most of the frequent switching of emotions within a stanza could not be reconstructed or justified. 

In poems, emotions are regularly mixed (already on line level) and are more interpretable within the whole poem. We therefore annotate lines hierarchically within the larger context of stanzas and the whole poem. Hence, we instruct the annotators to read a complete stanza or full poem, and then annotate each line in the context of its stanza. To reflect on the emotional complexity of poetry, we allow a maximum of two labels per line while avoiding heavy label fluctuations by encouraging annotators to reflect on their feelings to avoid `empty' annotations. Rather, they were advised to use fewer labels and more consistent annotation. This additional constraint is necessary to avoid ``wild'', non-reconstructable or non-justified annotations.

All subsequent batches (all except the first three) were only
annotated by two out of the three initial annotators, coincidentally those two who had the lowest initial agreement with each other. We asked these two experts to use the generated gold standard (48 poems;
majority votes of 3 annotators plus manual curation) as a reference
(``if in doubt, annotate according to the gold standard'').  This
eliminated some systematic differences between them\footnote{One
  person labeled lines with more negative emotions such as
  \textit{Uneasiness} and \textit{Annoyance} and the person labeled
  more positive emotions such as \textit{Vitality/Energy} and
  \textit{Beauty/Joy}.} and markedly improved the agreement levels,
roughly from 0.3--0.5~Cohen's $\kappa$ in the first three batches to around
0.6--0.8~$\kappa$ for all subsequent batches. This
annotation procedure relaxes the \textit{reader} perspective, as we encourage annotators (if in doubt) to annotate how they think the other annotators would annotate. However, we found that this
formulation improves the usability of the data and leads to a more
consistent annotation.

\begin{table}
\centering\small
\begin{tabularx}{\linewidth}{Xl}
\toprule
Factor & Items \\
\cmidrule(r){1-1}\cmidrule(l){2-2}
Negative emotions & anger/distasteful \\
Prototypical Aesthetic\par Emotions & beauty/sublime/being moved \\
Epistemic Emotions &interest/insight \\ 
Animation & motivation/inspiration \\ 
Nostalgia / Relaxation & nostalgic/calmed \\ 
Sadness & sad/melancholic \\ 
Amusement & funny/cheerful \\
\bottomrule
\end{tabularx}
\caption{Aesthetic Emotion Factors \protect\cite{schindler2017measuring}.}
\label{tab:factors}
\end{table}

\begin{figure*}[t]
\centering
\includegraphics[height=6.5cm,page=1]{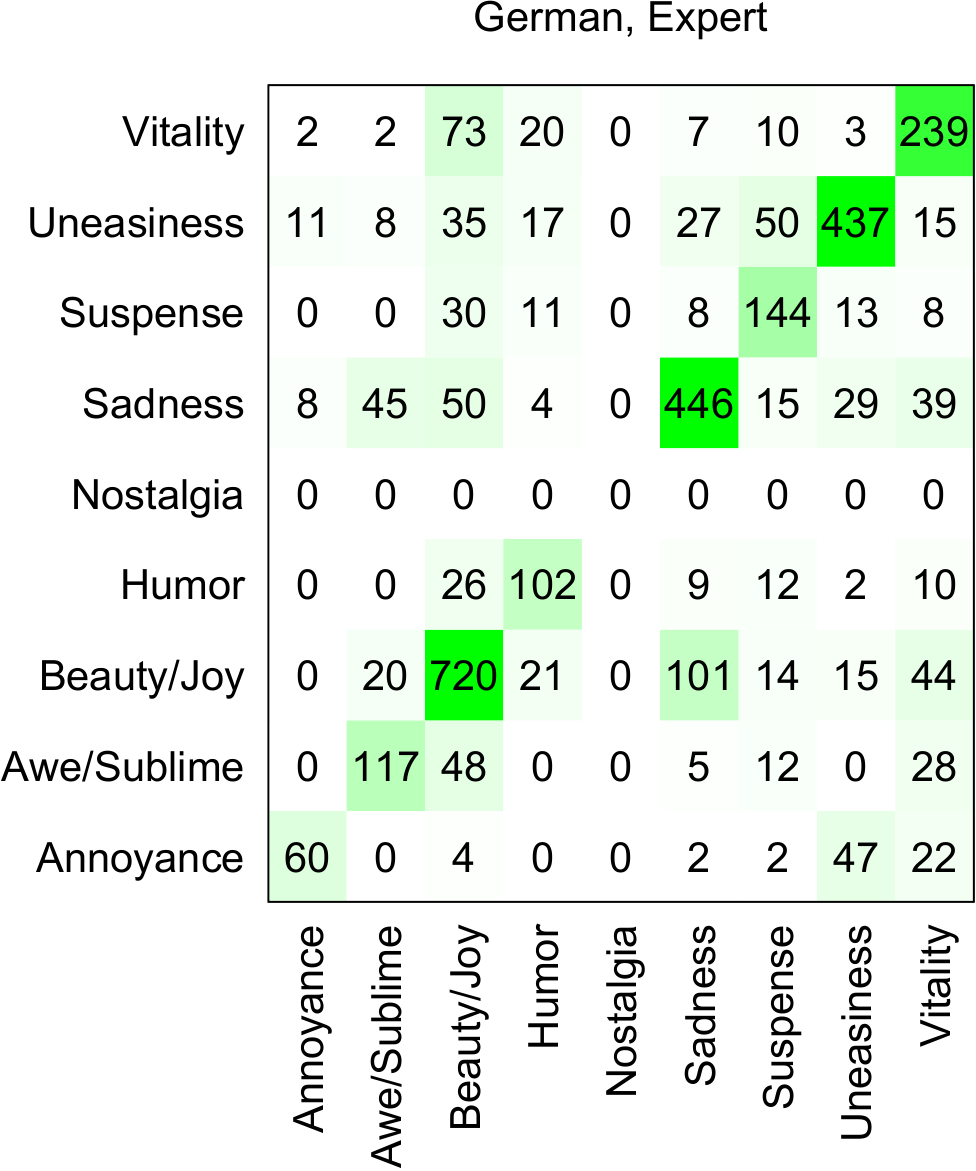}\ %
\includegraphics[height=6.5cm,page=2]{figures/heatmaps/heatmaps-crop}\ %
\includegraphics[height=6.5cm,page=3]{figures/heatmaps/heatmaps-crop}
\caption{Emotion cooccurrence matrices for the German and English expert annotation experiments and the English crowdsourcing experiment.}
\label{fig:heatmaps}
\end{figure*}

\subsection{Emotion Labels}\label{sec:emotionlabels}
We opt for
measuring the \textit{reader perspective} rather than the text surface or author's intent. To closer define and support conceptualizing our labels, we use particular `items', as they are used in psychological self-evaluations. These items consist of adjectives, verbs or short phrases. We build on top of
\newcite{schindler2017measuring} who proposed 43 items that were then grouped by a factor analysis based on self-evaluations of participants. The resulting factors are shown in Table \ref{tab:factors}. We attempt to cover all identified factors and supplement with basic emotions \cite{Ekman1992,Plutchik1991}, where possible.

\begin{table}
    \centering\small
    \setlength{\tabcolsep}{5pt}
    \begin{tabular}{lcc|cc|cc} \toprule
         & \multicolumn{2}{c}{\textbf{$\kappa$}} & \multicolumn{2}{c}{\textbf{Ann.~1 \%}} & \multicolumn{2}{c}{\textbf{Ann.~2 \%}} \\ 
         \cmidrule(r){2-3}\cmidrule(rl){4-5}\cmidrule(l){6-7}
         & en & de & en & de & en & de\\ \cmidrule(r){1-1}\cmidrule(rl){2-2}\cmidrule(rl){3-3}\cmidrule(rl){4-4}\cmidrule(rl){5-5}\cmidrule(rl){6-6}\cmidrule(rl){7-7}
        Beauty / Joy & .77 & .74 & .31 & .30 & .26 & .30\\
        Sadness & .72 & .77 & .21 & .20 &.20 & .18\\
        Uneasiness & .84 & .77 & .15 & .19 &.15 & .18\\
        Vitality / Energy & .50 & .63 & .12 & .11 &.18 & .13\\
        Awe / Sublime & .71 & .61 & .07 & .06 &.07 & .06\\
        Suspense & .58 & .65 & .04 & .07 &.07 & .08\\
        Humor & .81 & .68 & .04 & .05 &.04 & .05\\
        Nostalgia & .81 & --- & .03 & --- &.03 & ---\\
      Annoyance & .62 & .65&  .03 & .04 &.02 & .02\\
         \bottomrule
    \end{tabular}
    \caption{Cohen's kappa agreement levels and normalized line-level emotion frequencies for expert annotators (Nostalgia is not available in the German data).}
    \label{table:expert_agreements}
\end{table}

\begin{table}[t]
    \centering\small
    \begin{tabular}{lccc} \toprule
        & English & German \\ 
        \cmidrule(r){1-1}\cmidrule(lr){2-2}\cmidrule(l){3-3}
        avg. $\kappa$ & 0.707 & 0.688 \\
        \cmidrule(lr){2-2}\cmidrule(l){3-3}
        F1 & 0.775 & 0.774\\ 
       \cmidrule(lr){2-2}\cmidrule(l){3-3}
        F1 Majority & 0.323 & 0.323\\
        F1 Random & 0.108 & 0.119\\
         \bottomrule
    \end{tabular}
    \caption{Top: averaged kappa scores and micro-F1 agreement scores, taking one annotator as gold. Bottom: Baselines.}
    \label{table:expert_agreements_f1}
\end{table}

We started with a larger set of labels to then delete and substitute
(tone down) labels during the initial annotation process to avoid
infrequent classes and inconsistencies. Further, we conflate labels if
they show considerable confusion with each other.
These iterative improvements particularly affected
\textit{Confusion}, \textit{Boredom} and \textit{Other} that were very infrequently
annotated and had little agreement among annotators
($\kappa<.2$). For German, we also removed \textit{Nostalgia} 
($\kappa=.218$) after gold standard creation, but after consideration,
added it back for English, then achieving agreement. \textit{Nostalgia} is
still available in the gold standard (then with a second label
\textit{Beauty/Joy} or \textit{Sadness} to keep consistency). However,
\textit{Confusion}, \textit{Boredom} and \textit{Other} are not available in any
sub-corpus.

Our final set consists of nine classes, i.e., (in order of frequency)
\textit{Beauty/Joy, Sadness, Uneasiness, Vitality/Energy, Suspense, Awe/Sublime, Humor, Annoyance,} and \textit{Nostalgia}. In the
following, we describe the labels and give further details on the
aggregation process.

\textbf{Annoyance} (annoys me/angers me/felt frustrated): Annoyance
implies feeling annoyed, frustrated or even angry while reading the line/stanza.  We include the class \textit{Anger} here, as this was found to be too strong in intensity.

\textbf{Awe/Sublime} (found it overwhelming/sense of greatness):
\textit{Awe/Sublime} implies being overwhelmed by the line/stanza,
i.e., if one gets the impression of facing something sublime or if the
line/stanza inspires one with awe (or that the expression itself is sublime). Such emotions are often associated
with subjects like god, death, life, truth, etc. The term
\textit{Sublime} originated with
\newcite{kant2000critique} as one of the first aesthetic emotion
terms. \textit{Awe} is a more common English term.

\textbf{Beauty/Joy} (found it beautiful/pleasing/makes me
happy/joyful): \newcite{kant2000critique} already spoke of a ``feeling
of beauty'', and it should be noted that it is not a `merely pleasing
emotion'. 
Therefore, in our pilot annotations, \textit{Beauty} and
\textit{Joy} were separate labels. However,
\newcite{schindler2017measuring} found that items for \textit{Beauty}
and \textit{Joy} load into the same factors. Furthermore, our pilot
annotations revealed, while \textit{Beauty} is the more dominant and
frequent feeling, both labels regularly accompany each other, and they
often get confused across annotators. Therefore, we add \textit{Joy}
to form an inclusive label \textit{Beauty/Joy} that increases consistency.

\textbf{Humor} (found it funny/amusing): Implies feeling amused by
the line/stanza or if it makes one laugh.

\textbf{Nostalgia} (makes me nostalgic): Nostalgia is defined as a
sentimental longing for things, persons or situations in the past.  It
often carries both positive and negative feelings. However, since this
label is quite infrequent, and not available in all subsets of the
data, we annotated it with an additional \textit{Beauty/Joy} or
\textit{Sadness} label to ensure annotation consistency.

\textbf{Sadness} (makes me sad/touches me): If the line/stanza
makes one feel sad. It also includes a more general `being touched /
moved'.

\textbf{Suspense} (found it gripping/sparked my interest): Choose
\textit{Suspense} if the line/stanza keeps one in suspense (if it excites one or triggers one's curiosity).  We
removed \textit{Anticipation} from the earlier \textit{Suspense/Anticipation} label,
as \textit{Anticipation} appeared to us as being a more cognitive
prediction whereas Suspense is a far more straightforward emotion
item.

\textbf{Uneasiness} (found it ugly/unsettling/disturbing /
frightening/distasteful): This label covers situations when one
feels discomfort, when the line/stanza feels
distasteful/ugly, unsettling/disturbing or frightens one.  The labels
\textit{Ugliness} and \textit{Disgust} were conflated into
\textit{Uneasiness}, as both are seldom felt in poetry (being
inadequate/too strong/high in arousal), and typically lead to
\textit{Uneasiness}.

\textbf{Vitality/Energy} (found it invigorating/spurs me on/inspires me):
This label is meant for a line/stanza that has an inciting,
encouraging effect (if it conveys a feeling of movement,
energy and vitality which animates to action). Other terms: \textit{Animated}, \textit{Inspiration}, \textit{Stimulation} and \textit{Activation}.\footnote{Activation appears stable across cultures \cite{jackson2019emotion}}

\subsection{Agreement}

Table \ref{table:expert_agreements} shows the Cohen's $\kappa$
agreement scores among our two expert annotators for each emotion
category $e$ as follows. We assign each instance (a line in a poem) a
binary label indicating whether or not the annotator has annotated the
emotion category $e$ in question. From this, we obtain vectors
$v_i^e$, for annotators $i=0,1$, where each entry of $v_i^e$ holds the
binary value for the corresponding line. We then apply the $\kappa$
statistics to the two binary vectors $v_i^e$.  Additionally to
averaged $\kappa$, we report micro-F1 values in Table
\ref{table:expert_agreements_f1} between the multi-label annotations
of both expert annotators as well as the micro-F1 score of a random
baseline as well as of the majority emotion baseline (which labels
each line as \textit{Beauty/Joy}).

\begin{figure*}
\centering
\includegraphics[]{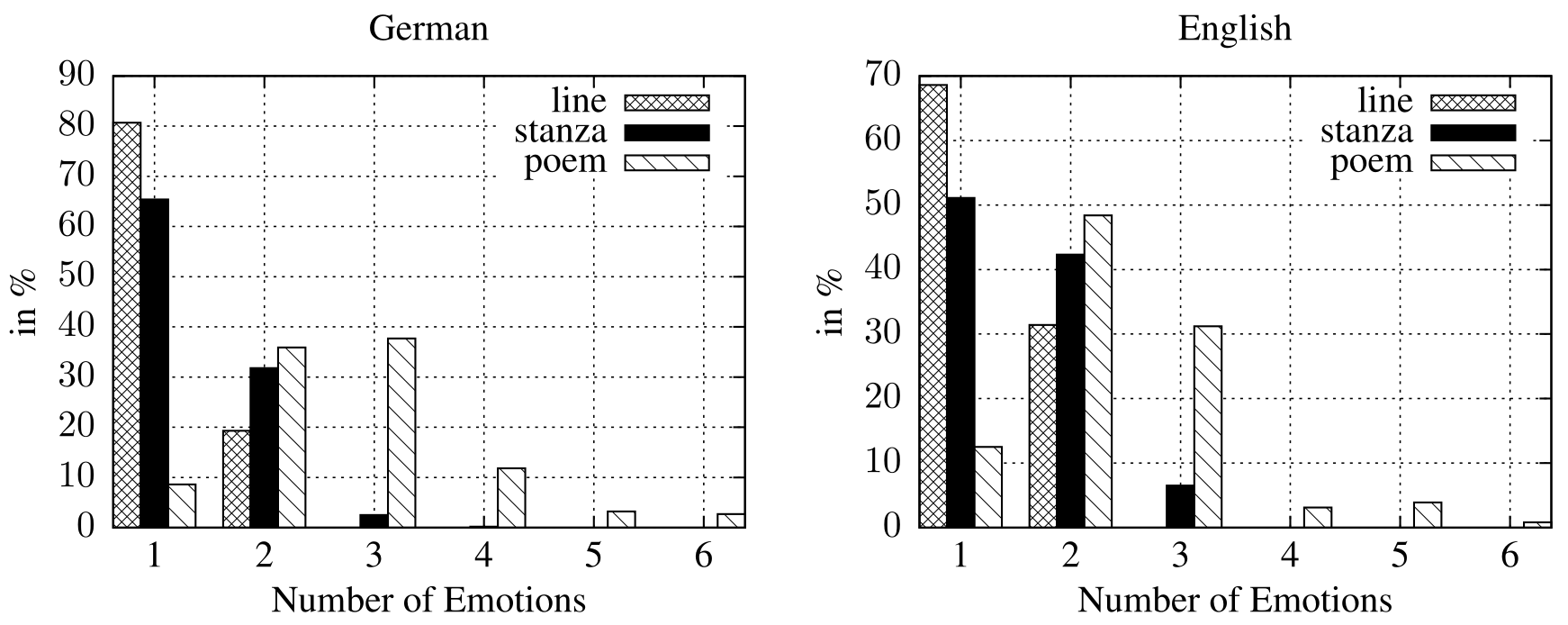}
  \caption{Distribution of number of distinct emotion labels per logical document level in the expert-based annotation. No whole poem has more than 6 emotions. No stanza has more than 4  emotions.}
  \label{fig:documentstructure}
\end{figure*}

We find that Cohen $\kappa$ agreement ranges from .84 for
\textit{Uneasiness} in the English data, .81 for \textit{Humor} and
\textit{Nostalgia}, down to German \textit{Suspense} (.65),
\textit{Awe/Sublime} (.61) and \textit{Vitality/Energy} for both languages (.50
English, .63 German).
Both annotators have a similar emotion frequency profile, where the
ranking is almost identical, especially for German. However, for
English, Annotator 2 annotates more \textit{Vitality/Energy} than
\textit{Uneasiness}.
Figure \ref{fig:heatmaps} shows the confusion matrices of labels
between annotators as heatmaps. Notably, \textit{Beauty/Joy} and
\textit{Sadness} are confused across annotators more often than other
labels. This is topical for poetry, and therefore not surprising: One
might argue that the beauty of beings and situations is only
beautiful because it is not enduring and therefore not to divorce from
the sadness of the vanishing of beauty \cite{benjamin2016goethes}.
We also find considerable confusion of \textit{Sadness} with
\textit{Awe/Sublime} and \textit{Vitality/Energy}, while the latter is also
regularly confused with \textit{Beauty/Joy}.

Furthermore, as shown in Figure \ref{fig:documentstructure}, we find
that no single poem aggregates to more than six emotion labels, while
no stanza aggregates to more than four emotion labels. However, most
lines and stanzas prefer one or two labels. German poems seem more
emotionally diverse where more poems have three labels than two
labels, while the majority of English poems have only two labels. This
is however attributable to the generally shorter English texts.

\section{Crowdsourcing Annotation}
After concluding the expert annotation, we performed a focused
crowdsourcing experiment, based on the final label set and items as
they are listed in Table \ref{tab:aggregation} and Section
\ref{sec:emotionlabels}. With this experiment, we aim to understand
whether it is possible to collect reliable judgements for aesthetic
perception of poetry from a crowdsourcing platform.  A second goal is
to see whether we can replicate the expensive expert annotations with
less costly crowd annotations.

We opted for a maximally simple annotation environment, where we asked
participants to annotate English 4-line stanzas with self-perceived
reader emotions. We choose English due to the higher availability of
English language annotators on crowdsourcing platforms. Each annotator
rates each stanza independently of surrounding context.

\subsection{Data and Setup}
\label{crowdguide}
For consistency and to simplify the task for the annotators, we opt
for a trade-off between completeness and granularity of the
annotation. Specifically, we subselect stanzas composed of four verses
from the corpus of 64 hand selected English poems. The resulting
selection of 59 stanzas is uploaded to Figure
Eight\footnote{\url{https://www.figure-eight.com/}} for annotation.

The annotators are asked to answer the following questions for each
instance.

\textbf{Question 1} (single-choice): Read the following stanza and
decide for yourself which emotions it evokes.

\textbf{Question 2} (multiple-choice): Which additional emotions does
the stanza evoke?

The answers to both questions correspond to the emotion labels we
defined to use in our annotation, as described in Section
\ref{sec:emotionlabels}. We add an additional answer choice ``None''
to Question 2 to allow annotators to say that a stanza does not evoke
any additional emotions.

Each instance is annotated by ten people. We restrict the task
geographically to the United Kingdom and Ireland and set the %internal
parameters on Figure Eight to only have the highest quality
annotators join the task. We pay \euro{0.09} per instance. The
final cost of the crowdsourcing experiment is \euro{74}.

\subsection{Results}\label{sec:crowd_results}

\begin{table*}
  \centering
    \begin{tabular}{lrrrrrrrrrrr}
    \toprule
          & \multicolumn{5}{c}{$\kappa$}          &       & \multicolumn{5}{c}{Counts} \\
\cmidrule{2-6}\cmidrule{8-12}    
Threshold & \multicolumn{1}{c}{$\geq1$} & \multicolumn{1}{c}{$\geq2$} & \multicolumn{1}{c}{$\geq3$} & \multicolumn{1}{c}{$\geq4$} & \multicolumn{1}{c}{$\geq5$} &       & \multicolumn{1}{c}{$\geq1$} & \multicolumn{1}{c}{$\geq2$} & \multicolumn{1}{c}{$\geq3$} & \multicolumn{1}{c}{$\geq4$} & \multicolumn{1}{c}{$\geq5$} \\
\cmidrule(r){1-1}\cmidrule{2-6}\cmidrule{8-12}
    Beauty / Joy & .21   & .41   & .46   & .28   & --    &       & 34.58 & 15.98 & 7.51  & 3.23  & 1.43 \\
    Sadness & .43   & .47   & .52   & .02   & $-$.04 &       & 43.34 & 28.99 & 17.77 & 9.52  & 2.82 \\
    Uneasiness & .18   & .25   & .08   & $-$.01 & --    &       & 36.47 & 16.33 & 5.49  & 1.54  & 1.04 \\
    Vitality & .15   & .26   & .19   & --    & --    &       & 25.62 & 7.34  & 2.02  & 1.05  & 1.00 \\
    Awe / Sublime & .31   & .17   & .37   & .46   & --    &       & 29.8  & 11.36 & 3.4   & 1.31  & 1.00 \\
    Suspense & .11   & .29   & .21   & .26   & --    &       & 39.12 & 17.8  & 6.54  & 1.97  & 1.04 \\
    Humor & .19   & .46   & .39   & $\approx$0 & --    &       & 19.26 & 5.36  & 2.1   & 1.22  & 1.07 \\
    Nostalgia & .23   & .01   & $-$.02 & --    & --    &       & 30.52 & 10.16 & 1.95  & 1.00   & 1.00 \\
    Annoyance & .01   & .07   & .66   & 0     & --    &       & 26.54 & 6.17  & 1.35  & 1.00   & 1.00 \\ 
    \cmidrule(r){1-1}\cmidrule{2-6}\cmidrule{8-12}
    Average & 0.20 & 0.27 & 0.32 & 0.14 &$-$0.04 &&31.69 & 13.28 & 5.35 & 2.43 & 1.27   \\
    \bottomrule
    \end{tabular}%
     \caption{Results obtained via boostrapping for annotation aggregation. The row \emph{Threshold} shows how many people within a group of five annotators should agree on a particular emotion. The column labeled \emph{Counts} shows the average number of times certain emotion was assigned to a stanza given the threshold. Cells with `--' mean that neither of two groups satisfied the threshold.}
  \label{tab:aggregation}%
\end{table*}%

In the following, we determine the best aggregation strategy regarding
the 10 annotators with bootstrap resampling. For instance, one could
assign the label of a specific emotion to an instance if just one
annotators picks it, or one could assign the label only if all
annotators agree on this emotion. To evaluate this, we repeatedly pick
two sets of 5 annotators each out of the 10 annotators for each of the 59 stanzas, 1000 times
overall (i.e., 1000$\times$59 times, bootstrap resampling).  For each
of these repetitions, we compare the agreement of these two groups of
5 annotators. Each group gets assigned with an adjudicated emotion
which is accepted if at least one annotator picks it, at least two 
annotators pick it, etc.\ up to all five pick it.

We show the results in Table~\ref{tab:aggregation}. The $\kappa$
scores show the average agreement between the two groups of five
annotators, when the adjudicated class is picked based on the
particular threshold of annotators with the same label choice. We see
that some emotions tend to have higher agreement scores than others,
namely \emph{Annoyance} (.66), \emph{Sadness} (up to .52), and
\emph{Awe/Sublime}, \emph{Beauty/Joy}, \emph{Humor} (all .46). The
maximum agreement is reached mostly with a threshold of 2 (4 times) or
3 (3 times).

We further show in the same table the average numbers of labels from
each strategy. Obviously, a lower threshold leads to higher numbers
(corresponding to a disjunction of annotations for each emotion). The
drop in label counts is comparably drastic, with on average 18 labels
per class. Overall, the best average $\kappa$ agreement (.32) is less
than half of what we saw for the expert annotators (roughly
.70). Crowds especially disagree on many more intricate emotion labels
(Uneasiness, Vitality/Energy, Nostalgia, Suspense).

We visualize how often two emotions are used to label an instance in a
confusion table in Figure~\ref{fig:heatmaps}. Sadness is used most
often to annotate a stanza, and it is often confused with Suspense,
Uneasiness, and Nostalgia. Further, Beauty/Joy partially overlaps with
Awe/Sublime, Nostalgia, and Sadness.

On average, each crowd annotator uses two emotion labels per stanza
(56\% of cases); only in 36\% of the cases the annotators use one
label, and in 6\% and 1\% of the cases three and four labels,
respectively.  This contrasts with the expert annotators, who use one
label in about 70\% of the cases and two labels in 30\% of the cases
for the same 59 four-liners.  Concerning frequency distribution for
emotion labels, both experts and crowds name Sadness and Beauty/Joy as
the most frequent emotions (for the `best' threshold of 3) and
Nostalgia as one of the least frequent emotions. The Spearman rank
correlation between experts and crowds is about 0.55 with respect to
the label frequency distribution, indicating that crowds could
replace experts to a moderate degree when it comes to extracting,
e.g., emotion distributions for an author or time period. Now, we
further compare crowds and experts in terms of whether crowds could
replicate expert annotations also on a finer stanza level (rather than
only on a distributional level).

\subsection{Comparing Experts with Crowds}

To gauge the quality of the crowd annotations in comparison with our
experts, we calculate agreement on the emotions between experts and an
increasing group size from the crowd.  For each stanza instance $s$,
we pick $N$ crowd workers, where $N\in\{4,6,8,10\}$, then pick their
majority emotion for $s$, and additionally pick their second ranked
majority emotion if at least $\frac{N}{2}-1$ workers have chosen
it.\footnote{For workers, we additionally require that an emotion has
  been chosen by at least $2$ workers.} For the experts, we aggregate
their emotion labels on stanza level, then perform the same strategy
for selection of emotion labels.  Thus, for $s$, both crowds and
experts have 1 or 2 emotions.  For each emotion, we then compute
Cohen's $\kappa$ as before. Note that, compared to our previous
experiments in Section~\ref{sec:crowd_results} with a threshold, each
stanza now receives an emotion annotation (exactly one or two emotion
labels), both by the experts and the crowd-workers.

\begin{figure}
    \centering
    \includegraphics[]{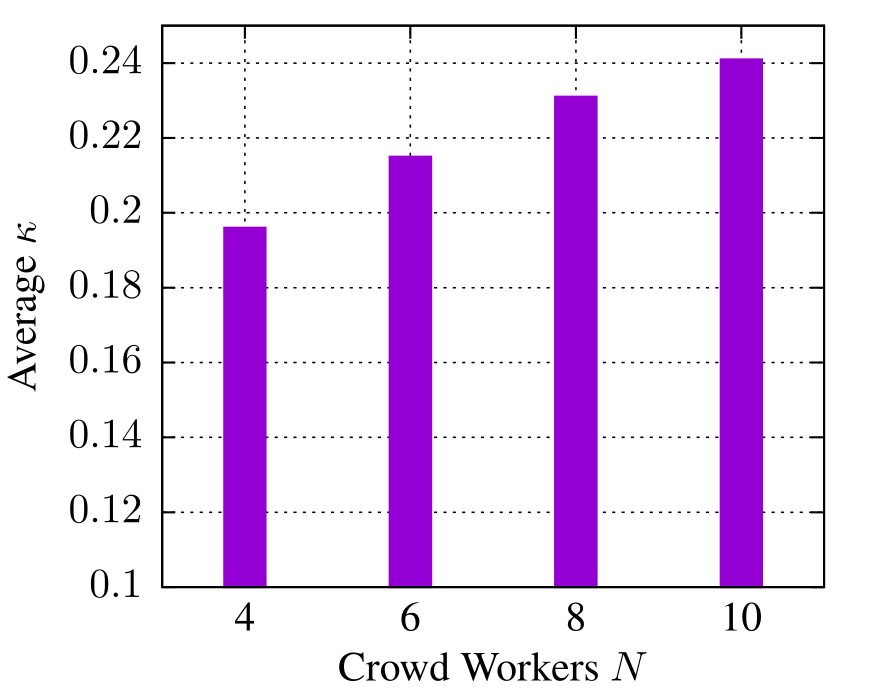}
    \caption{Agreement between experts and crowds as a function of the number $N$ of crowd workers.}
    \label{fig:expert_vs_crowds}
\end{figure}

In Figure \ref{fig:expert_vs_crowds}, we plot agreement between
experts and crowds on stanza level as we vary the number $N$ of crowd
workers involved.  On average, there is roughly a steady linear
increase in agreement as $N$ grows, which may indicate that $N=20$ or
$N=30$ would still lead to better agreement. Concerning individual
emotions, \emph{Nostalgia} is the emotion with the least agreement, as
opposed to \emph{Sadness} (in our sample of 59 four-liners): the
agreement for this emotion grows from $.47$ $\kappa$ with $N=4$ to
$.65$ $\kappa$ with $N=10$. \emph{Sadness} is also the most frequent
emotion, both according to experts and crowds. Other emotions for
which a reasonable agreement is achieved are \emph{Annoyance,
  Awe/Sublime, Beauty/Joy, Humor} ($\kappa$ > 0.2). Emotions with
little agreement are \emph{Vitality/Energy, Uneasiness, Suspense, Nostalgia}
($\kappa$ < 0.2).

By and large, we note from Figure~\ref{fig:heatmaps} that expert
annotation is more restrictive, with experts agreeing more often on
particular emotion labels (seen in the darker diagonal).

The results
of the crowdsourcing experiment, on the other hand, are a mixed bag as
evidenced by a much sparser distribution of emotion labels. However,
we note that these differences can be caused by 1)~the disparate
training procedure for the experts and crowds, and 2)~the lack of
opportunities for close supervision and on-going training of the
crowds, as opposed to the in-house expert annotators.

In general, however, we find that substituting experts with crowds is
possible to a certain degree. Even though the crowds' labels look
inconsistent at first view, there appears to be a good signal in their
\emph{aggregated} annotations, helping to approximate expert
annotations to a certain degree. The average $\kappa$ agreement (with
the experts) we get from $N=10$ crowd workers (0.24) is still
considerably below the agreement among the experts (0.70).

\section{Modeling}
To estimate the difficulty of automatic classification of our
data set, we perform multi-label\footnote{We found that single-label classification had only marginally better performance, even though the task is simpler.} document classification (of stanzas) with BERT
\cite{Devlin2019}. For this experiment we aggregate all labels for a stanza and sort them by frequency, both for the gold standard and the raw expert annotations.
As can be seen in Figure \ref{fig:documentstructure}, a stanza bears a minimum of one and a maximum of four emotions.
 Unfortunately, the label \textit{Nostalgia} is only available 16 times in the German data (the gold standard) as a second label (as discussed in Section \ref{sec:emotionlabels}). None of our models was able to learn this label for German. Therefore we omit it, leaving us with eight proper labels.

We use the code and the pre-trained BERT models of
\textsc{Farm},\footnote{\url{https://github.com/deepset-ai/FARM}} provided by \url{deepset.ai}. 
We test the multilingual-uncased model (\textsc{Multiling}),
the german-base-cased model (\textsc{Base}),\footnote{There was no uncased model available.} the german-dbmdz-uncased model
(\textsc{Dbmdz}),\footnote{\url{https://github.com/dbmdz} a model by the
  Bavarian state library that was also trained on literature.} and we tune the \textsc{Base} model on 80k stanzas of the German Poetry Corpus DLK \cite{haider2019semantic} for 2 epochs, both on token (masked words) and sequence (next line) prediction (\textsc{Base}$_{\textsc{Tuned}}$).

We split the randomized German dataset so that each label is at least 10 times in the validation set (63 instances, 113 labels), and at least 10 times in the test set (56 instances, 108 labels) and leave the rest for training (617 instances, 946 labels).\footnote{We do the same for the English data (at least 5 labels) and add the stanzas to the respective sets.} We train BERT for 10 epochs (with a batch size of 8), % and 
%evaluate every 20 batches,%\footnote{An epoch iterates over $\frac{617 instances}{8 batch size}$ = 78 batches}
optimize with entropy loss, and report F1-micro on the test set.
See Table \ref{tab:bert} for the results.

We find that the multilingual model cannot handle infrequent
categories, i.e., \textit{Awe/Sublime, Suspense} and
\textit{Humor}. However, increasing the dataset with English data
improves the results, suggesting that the classification would largely benefit from more annotated data. The best model overall is \textsc{DBMDZ} (.520), showing a balanced response on both validation and test set. See Table \ref{tab:classifyallemotions} for a breakdown of all emotions as predicted by the this model. Precision is mostly higher than recall. The labels \textit{Awe/Sublime}, \textit{Suspense} and \textit{Humor} are harder to predict than the other labels. 

The \textsc{BASE} and \textsc{BASE}$_{\textsc{TUNED}}$ models perform slightly worse than \textsc{DBMDZ}.
The effect of tuning of the \textsc{BASE} model is questionable, probably because of the restricted vocabulary (30k). We found that tuning on poetry does not show obvious improvements. Lastly, we find that models that were trained on lines (instead of stanzas) do not achieve the same F1 (\textasciitilde.42 for the German models).

%\begin{table}[t]
%\begin{small}
%    \centering
%    \begin{tabular}{lcccc}
%    \toprule
%     & \multicolumn{2}{c}{German} &  %\multicolumn{2}{c}{Multiling.} \\
%     \cmidrule(lr){2-3}\cmidrule(l){4-5}
%     Model   &  dev  &   test    &   dev &   test    \\
%     \cmidrule(r){1-1}\cmidrule(rl){2-3}\cmidrule(l){4-5}
%    Majority (Beauty)       &  .318   &   .285 %                 &   .318    &     .232      \\
%    MULTILING   & .333  & .416            &   %.431   &    .452       \\
%        DEEPSET        &   .565    &     \textbf{.571}                 &  --     &    -- %      \\
%    DBMDZ        &   \textbf{.609}    &     .507     &        --     &      --     \\
%    $LARGE_{en}$       &   -    &    -       & %     &           &       &           \\
%    TUNED$_{\text{DEEPSET}}$ &  .551     &      .491               &   --    &    --       \\
%    TUNED\_MULTI&       &           &       & %          &       &           \\
%    \bottomrule
%    \end{tabular}
%    \caption{BERT-based classification on stanza-level, single-label}
%    \label{tab:bert}
%    \end{small}
%\end{table}

\begin{table}[t]
\begin{small}
    \centering
    \begin{tabular}{lcccc}
    \toprule
     & \multicolumn{2}{c}{German} &  \multicolumn{2}{c}{Multiling.} \\
     \cmidrule(lr){2-3}\cmidrule(l){4-5}
     Model   &  dev  &   test    &   dev &   test    \\
     \cmidrule(r){1-1}\cmidrule(rl){2-3}\cmidrule(l){4-5}
    Majority       &  .212   &   .167                  &   .176    &     .150      \\
    MULTILING   & .409  & .341            &   .461   &    .384       \\
        BASE        &   .500    &     .439                 &  --     &    --       \\
%    $LARGE_{en}$       &   -    &    -       &      &           &       &           \\
    BASE$_{\text{TUNED}}$ &  .477     &      .514               &   --    &    --       \\
        DBMDZ        &   .520    &     \textbf{.520}     &        --     &      --     \\
%    TUNED\_MULTI&       &           &       &           &       &           \\
    \bottomrule
    \end{tabular}
    \caption{BERT-based multi-label classification on stanza-level.}
    \label{tab:bert}
    \end{small}
\end{table}

\begin{table}[t]
  \begin{small}
    \centering
    \begin{tabular}{llllr}
      \toprule 
      Label & Precision    & Recall & F1 & Support \\
      \cmidrule(r){1-1}\cmidrule(rl){2-2}\cmidrule(rl){3-3}\cmidrule(lr){4-4}\cmidrule(l){5-5}
  Beauty/Joy  &   0.5000  &  0.5556  &  0.5263  &      18\\
     Sadness   &  0.5833  &  0.4667  &  0.5185  &      15\\
  Uneasiness&     0.6923  &  0.5625  &  0.6207  &      16\\
    Vitality/Energy &    1.0000  &  0.5333  &  0.6957  &      15\\
   Annoyance  &   1.0000  &  0.4000  &  0.5714  &      10\\
 Awe/Sublime   &  0.5000  &  0.3000  &  0.3750  &      10\\
    Suspense    & 0.6667 &   0.1667  &  0.2667  &      12\\
       Humor    & 1.0000&    0.2500  &  0.4000  &      12\\
  \cmidrule(r){1-1}\cmidrule(rl){2-2}\cmidrule(rl){3-3}\cmidrule(rl){4-4}\cmidrule(l){5-5}
   micro avg    & 0.6667 &   0.4259  &  0.5198      & 108\\
   macro avg    & 0.7428  &  0.4043  &  0.4968 &      108\\
weighted avg    & 0.7299  &  0.4259  &  0.5100  &     108\\
 samples avg    & 0.5804  &  0.4464 &   0.4827   &    108\\
   \bottomrule
    \end{tabular}
    \caption{Recall and precision scores of the best model (dbmdz) for each emotion on the test set. `Support': number of instances with this label.}
    \label{tab:classifyallemotions}
  \end{small}
\end{table}

%\subsubsection{Lines}

%\begin{table}[!ht]
%\begin{small}
%    \centering
%    \begin{tabular}{l|c|c|c|c|c|c|}
%     Language $\rightarrow$      & \multicolumn{2}{c|}{German} & \multicolumn{2}{c|}{English} & \multicolumn{2}{c|}{Multiling.} \\
%     Model  $\downarrow$         &  dev  &   test    &   dev &   test    &  dev  &   test    \\
%                \hline
%    Majority (Beauty)       &  .318   &   .285        &       &           &       &           \\
%    $MULTILING$   &   &  &       &           &       &           \\
%        $DEEPSET_{de}$        &       &       &       &           &       &           \\
%    $DBMDZ_{de}$        &   .45   &           &       &           &       &           \\
%    $LARGE_{en}$       &   -    &    -       &      &           &       &           \\
%    $TUNED_{DEEPSET}$ &       &         &       &           &       &           \\
%    TUNED\_MULTI&       &           &       &           &       &           \\
%    \hline
%    \end{tabular}
%    \caption{Accuracy (F1micro). BERT Document classification on lines}
%    \label{tab:my_label}
%    \end{small}
%\end{table}

%\subsubsection{Sentence Transformer \& MLP}
%To obtain a representation for our lines, we use average sentence embeddings from BERT \cite{reimers2019sentence}.
%\url{https://github.com/UKPLab/sentence-transformers/blob/master/README.md}

%Sentence Transformer BERT + MLP (75\% F1 for English). 
%MLP from sklearn \url{https://scikit-learn.org/stable/modules/generated/sklearn.neural_network.MLPClassifier.html}

%\subsection{Sequence Prediction}
%(BiLSTM + CNN + CRF) $\rightarrow$ Maybe move this to a COLING/ACL paper
%\url{https://github.com/UKPLab/emnlp2017-bilstm-cnn-crf}
\section{Concluding Remarks}
In this paper, we presented a dataset of German and English poetry
annotated with reader response to reading poetry.
We argued that basic emotions as proposed by psychologists (such as Ekman and Plutchik) that are often used in emotion analysis from text are of little use for the annotation of poetry reception. We instead conceptualized aesthetic emotion labels and showed that a closely supervised annotation task results in substantial agreement---in terms of $\kappa$ score---on the final dataset.

The task of collecting reader-perceived emotion response to poetry in
a crowdsourcing setting is not straightforward. In contrast to expert
annotators, who were closely supervised and reflected upon the task,
the annotators on crowdsourcing platforms are difficult to control and
may lack necessary background knowledge to perform the task at
hand. However, using a larger number of crowd annotators may lead to
finding an aggregation strategy with a better trade-off between quality
and quantity of adjudicated labels. For future work, we thus propose to
repeat the experiment with larger number of crowdworkers, and develop
an improved training strategy that would suit the crowdsourcing
environment.

The dataset presented in this paper can be of use for different
application scenarios, including multi-label emotion classification,
style-conditioned poetry generation, investigating the influence of
rhythm/prosodic features on emotion, or analysis of authors, genres and diachronic
variation (e.g., how emotions are represented differently in certain
periods).

Further, though our modeling experiments are still rudimentary, we
propose that this data set can be used to investigate the intra-poem
relations either through multi-task learning
\cite{schulz-etal-2018-multi} and/or with the help of hierarchical
sequence classification approaches.

\section*{Acknowledgements}
A special thanks goes to Gesine Fuhrmann, who created the guidelines
and tirelessly documented the annotation progress. Also thanks to
Annika Palm and Debby Trzeciak who annotated and gave lively
feedback. For help with the conceptualization of labels we thank Ines
Schindler. This research has been partially conducted within the CRETA
center (\url{http://www.creta.uni-stuttgart.de/}) which is funded by
the German Ministry for Education and Research (BMBF) and partially
funded by the German Research Council (DFG), projects SEAT (Structured
Multi-Domain Emotion Analysis from Text, KL 2869/1-1). This work has also been supported by the German Research Foundation as part of the Research Training Group Adaptive Preparation of Information from Heterogeneous Sources (AIPHES) at the Technische Universität Darmstadt under grant No. GRK 1994/1.

\vfill

\section*{Appendix}

We illustrate two examples of our German gold standard annotation, a poem each by Friedrich Hölderlin and Georg
Trakl, and an English poem by Walt Whitman. Hölderlin's text stands out, because the mood changes starkly
from the first stanza to the second, from \textit{Beauty/Joy} to
\textit{Sadness}.  Trakl's text is a bit more complex with bits of
\textit{Nostalgia} and, most importantly, a mixture of
\textit{Uneasiness} with \textit{Awe/Sublime}. Whitman's poem is an example of \textit{Vitality} and its mixing with \textit{Sadness}. The English annotation was unified by us for space constraints. For the full annotation please see \url{https://github.com/tnhaider/poetry-emotion/}

\subsection*{Friedrich Hölderlin: Hälfte des Lebens (1804)}

\begin{scriptsize}
\centering
\setlength{\tabcolsep}{2pt}
\begin{tabular}{ll}
Mit gelben Birnen hänget & \textcolor{magenta}{[Beauty/Joy]} \\
Und voll mit wilden Rosen & \textcolor{magenta}{[Beauty/Joy]}  \\
Das Land in den See, & \textcolor{magenta}{[Beauty/Joy]} \\
Ihr holden Schwäne, & \textcolor{magenta}{[Beauty/Joy]} \\
Und trunken von Küssen & \textcolor{magenta}{[Beauty/Joy]} \\
Tunkt ihr das Haupt & \textcolor{magenta}{[Beauty/Joy]} \\
Ins heilignüchterne Wasser. & \textcolor{magenta}{[Beauty/Joy]}\\
 & \\
Weh mir, wo nehm' ich, wenn & \textcolor{blue}{[Sadness]} \\
Es Winter ist, die Blumen, und wo & \textcolor{blue}{[Sadness]}\\
Den Sonnenschein, & \textcolor{blue}{[Sadness]}\\
Und Schatten der Erde? & \textcolor{blue}{[Sadness]}\\
Die Mauern stehn & \textcolor{blue}{[Sadness]}\\
Sprachlos und kalt, im Winde & \textcolor{blue}{[Sadness]}\\
Klirren die Fahnen. & \textcolor{blue}{[Sadness]}\\
\end{tabular}
\end{scriptsize}

\subsection*{Georg Trakl: In den Nachmittag geflüstert (1912)}

\begin{scriptsize}
\centering
\setlength{\tabcolsep}{2pt}
\begin{tabular}{ll}
Sonne, herbstlich dünn und zag, & \textcolor{magenta}{[Beauty/Joy]} \textcolor{green}{[Nostalgia]}\\
Und das Obst fällt von den Bäumen. & \textcolor{magenta}{[Beauty/Joy]} \textcolor{green}{[Nostalgia]}\\
Stille wohnt in blauen Räumen & \textcolor{magenta}{[Beauty/Joy]}\\
Einen langen Nachmittag. & \textcolor{magenta}{[Beauty/Joy]}\\
  & \\
Sterbeklänge von Metall; & \textcolor{blue}{[Sadness]} \textcolor{olive}{[Uneasiness]}\\
Und ein weißes Tier bricht nieder. & \textcolor{blue}{[Sadness]} \textcolor{olive}{[Uneasiness]} \\
Brauner Mädchen rauhe Lieder & \textcolor{blue}{[Sadness]} \textcolor{green}{[Nostalgia]}\\
Sind verweht im Blätterfall. & \textcolor{blue}{[Sadness]} \textcolor{green}{[Nostalgia]}\\
 & \\
Stirne Gottes Farben träumt, & \textcolor{olive}{[Uneasiness]} \textcolor{orange}{[Awe/Sublime]} \\
Spürt des Wahnsinns sanfte Flügel. & \textcolor{olive}{[Uneasiness]} \textcolor{orange}{[Awe/Sublime]} \\
Schatten drehen sich am Hügel &  \textcolor{olive}{[Uneasiness]} \textcolor{orange}{[Awe/Sublime]} \\
Von Verwesung schwarz umsäumt. &  \textcolor{olive}{[Uneasiness]} \textcolor{orange}{[Awe/Sublime]} \\
  & \\
Dämmerung voll Ruh und Wein; & \textcolor{magenta}{[Beauty/Joy]}\\
Traurige Guitarren rinnen. & \textcolor{magenta}{[Beauty/Joy]}\\
Und zur milden Lampe drinnen & \textcolor{magenta}{[Beauty/Joy]}\\
Kehrst du wie im Traume ein. & \textcolor{magenta}{[Beauty/Joy]}\\
\end{tabular}
\end{scriptsize}

\subsection*{Walt Whitman: O Captain! My Captain! (1865)	}
\begin{scriptsize}
\centering
\setlength{\tabcolsep}{2pt}
\begin{tabular}{ll}
O Captain! my Captain! our fearful trip is done, &\textcolor{magenta}{[Beauty/Joy]}\\
The ship has weather'd every rack, the prize we sought is won,&	\textcolor{magenta}{[Beauty/Joy]}\\
The port is near, the bells I hear, the people all exulting,&	\textcolor{magenta}{[Beauty/Joy]}\\
While follow eyes the steady keel, the vessel grim and daring;&	\textcolor{magenta}{[Beauty/Joy]}\\
But O heart! heart! heart! &	\textcolor{blue}{[Sadness]}	\\
O the bleeding drops of red,&	\textcolor{blue}{[Sadness]}	\\
Where on the deck my Captain lies,&	\textcolor{blue}{[Sadness]}	\\
Fallen cold and dead.&	\textcolor{blue}{[Sadness]}	\\
\\
O Captain! my Captain! rise up and hear the bells; &	\textcolor{orange}{[Vitality]}	\\
Rise up -- for you the flag is flung -- for you the bugle trills,&	\textcolor{orange}{[Vitality]}	\\
For you bouquets and ribbon'd wreaths -- \\
\hspace{3cm}for you the shores a-crowding,	&\textcolor{orange}{[Vitality]}	\\
For you they call, the swaying mass, their eager faces  turning;&	\textcolor{orange}{[Vitality]}	\\
Here Captain! dear father! & \textcolor{orange}{[Vitality]}	\\
This arm beneath your head!	&\textcolor{orange}{[Vitality]}	\\
It is some dream that on the deck,	&\textcolor{blue}{[Sadness]}		\\
You've fallen cold and dead.&	\textcolor{blue}{[Sadness]}		\\
\\
My Captain does not answer, his lips are pale and still,&	\textcolor{blue}{[Sadness]}		\\
My father does not feel my arm, he has no pulse nor will,&	\textcolor{blue}{[Sadness]}		\\
The ship is anchor'd safe and sound, its voyage closed and done,&	\textcolor{orange}{[Vitality]}	\textcolor{blue}{[Sadn.]}\\
From fearful trip the victor ship comes in with object won;&	\textcolor{orange}{[Vitality]}	\textcolor{blue}{[Sadn.]}	 \\
Exult O shores, and ring O bells!&	\textcolor{orange}{[Vitality]}	\textcolor{blue}{[Sadn.]}	\\
But I with mournful tread,&	\textcolor{blue}{[Sadness]}		\\
Walk the deck my Captain lies,&	\textcolor{blue}{[Sadness]}		\\
Fallen cold and dead.&	\textcolor{blue}{[Sadness]}		\\
\end{tabular}
\end{scriptsize}

\section{Bibliographical References}
\label{main:ref}

\bibliographystyle{lrec}
\bibliography{biblio}

\end{document}